\documentclass[10pt,twocolumn,letterpaper]{article}

\usepackage{cvpr}              

\usepackage[dvipsnames,table]{xcolor}

\usepackage{graphicx}
\usepackage{amsmath}
\usepackage{amssymb}
\usepackage{booktabs}
\usepackage{multirow}
\usepackage{pifont}
\newcommand{\cmark}{\ding{51}}%
\newcommand{\xmark}{\ding{55}}%
\usepackage{multirow}
\newcommand*{\myalign}[2]{\multicolumn{1}{#1}{#2}}

\usepackage{braket}
\usepackage{cellspace}
\usepackage{float} 
\usepackage{mathtools}

\definecolor{cvprblue}{rgb}{0.21,0.49,0.74}
\usepackage[pagebackref,breaklinks,colorlinks,citecolor=cvprblue]{hyperref}

\def\paperID{1480} 
\def\confName{CVPR}
\def\confYear{2024}

\definecolor{mygrey}{RGB}{230,230,230} 
\definecolor{myblue}{RGB}{230,230,255}
\definecolor{myred}{RGB}{251,231,231}
\definecolor{myyellow}{RGB}{255,255,175}

\definecolor{tickgreen}{RGB}{127,201,127}
\definecolor{crossred}{RGB}{240,2,127}

\DeclarePairedDelimiter\abs{\lvert}{\rvert}%
\newcommand{\lpnorm}[1]{\left\lVert #1 \right\rVert}

\newcommand*{\belowrulesepcolor}[1]{%
  \noalign{%
    \kern-\belowrulesep 
    \begingroup 
      \color{#1}%
      \hrule height\belowrulesep 
    \endgroup 
  }%
} 
\newcommand*{\aboverulesepcolor}[1]{%
  \noalign{%
    \begingroup 
      \color{#1}%
      \hrule height\aboverulesep 
    \endgroup 
    \kern-\aboverulesep 
  }%
}

\def \alambicdeit {\includegraphics[width=0.025\linewidth]{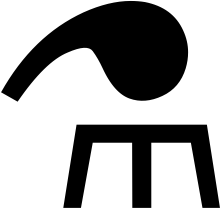}\xspace}

\def \alambicdeitcol {\includegraphics[width=0.07\linewidth]{figures/alembic-crop.pdf}\xspace}

\def \fire {\includegraphics[width=0.015\linewidth]{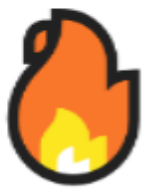}\xspace}

\def \ice {\includegraphics[width=0.015\linewidth]{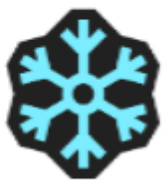}\xspace}

\begin{document}

\title{$V_kD:$ Improving Knowledge Distillation using Orthogonal Projections}

\author{Roy Miles
\hspace{6pt}
Ismail Elezi
\hspace{6pt}
Jiankang Deng
\vspace{.6ex}
\\
Huawei Noah's Ark Lab
}

\maketitle

\begin{abstract}
   Knowledge distillation is an effective method for training small and efficient deep learning models. However, the efficacy of a single method can degenerate when transferring to other tasks, modalities, or even other architectures. To address this limitation, we propose a novel constrained feature distillation method.
   This method is derived from a small set of core principles, which results in two emerging components: an orthogonal projection and a task-specific normalisation. Equipped with both of these components, our transformer models can outperform all previous methods on ImageNet and reach up to a 4.4\% relative improvement over the previous state-of-the-art methods. To further demonstrate the generality of our method, we apply it to object detection and image generation, whereby we obtain consistent and substantial performance improvements over state-of-the-art. Code and models are publicly available\footnote{\url{https://github.com/roymiles/vkd}}.
\end{abstract}


\section{Introduction}
\label{sec:intro}

Deep learning has achieved remarkable success across a wide variety of tasks in computer vision~\cite{Krizhevsky2012ImageNetNetworks}, audio~\cite{deep_learning_audio_signal_processing}, and language~\cite{devlin2018bert} domains. However, its adoption is often coupled with increasing computational costs which has limited its application on resource constrained devices such as mobile phones. Fortunately, there have been many techniques proposed to train fast and efficient networks, such as weight pruning~\cite{Lecun1990OptimalDamage, Lin2020HRank:Map, Guo2020DMCP:Networks}, quantisation~\cite{Bulat2019XNOR-Net++:Networks, Zhou2016DoReFa-Net:Gradients}, and knowledge distillation~\cite{Hinton2015DistillingNetwork, Tian2019ContrastiveDistillation, Miles2020CascadedSelf-distillation, Chen2021DistillingReview}.

Knowledge distillation (KD) in particular has shown great success \cite{Tian2019ContrastiveDistillation, Miles2023MobileVOS:Distillation, Fang2021ContrastiveDistillationv2, bhardwaj2019dream, lopezpaz2016unifying}.
Its main idea is to utilise the pre-trained knowledge of a much larger (teacher) model to supervise the training of a much smaller (student) model. 
Traditional KD~\cite{Hinton2015DistillingNetwork, Beyer2022KnowledgeConsistent} methods have focused on image classification by using the softmax predictions of the teacher as ground-truth labels for the student.
However, doing so has made them limited and only applicable to specific modalities, or tasks. 
Although feature distillation~\cite{Tian2019ContrastiveDistillation,Chen2020WassersteinDistillation} can relax this constraint on the downstream task or modality, its adoption can incur significant computational costs due to the construction of expensive relational objects~\cite{Park2019RelationalDistillation, Miles2022InformationDistillation,Miles2023MobileVOS:Distillation} and memory banks~\cite{Tian2019ContrastiveDistillation}. 
Most feature distillation pipelines can be described as using some projection~\cite{Romero2015FitNets:Nets}, alignment~\cite{chen2022dearkd}, or fusion module~\cite{Chen2021DistillingReview}. These components, though widely used, tend to rely on heuristic design choices. Such heuristic-driven approaches often fall short in delivering new insights into the underlying mechanics of distillation and struggle to adapt to diverse tasks without the introduction of several auxiliary losses.~\cite{Liu2019StructuredSegmentation, cui2023kddlgan, zhang2022wavelet, Chen2021DistillingReview}. Furthermore, these additional losses introduce additional hyperparameters which will require tuning for specific tasks or settings.

\begin{figure}
\centering
\resizebox{1.\linewidth}{!}{%
\includegraphics{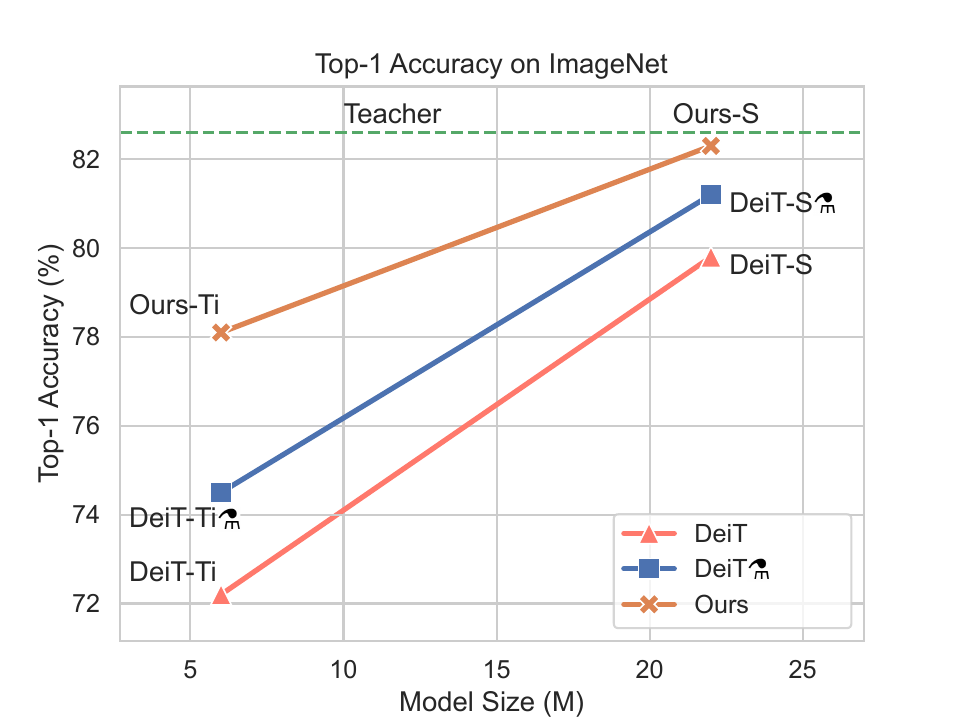}
}
\caption{Comparison to both DeiT and DeiT\alambicdeit~\cite{Touvron2021TrainingAttention} on ImageNet-1K, where DeiT\alambicdeit is a distilled DeiT model using a distillation token. Our proposed distillation method achieves significant improvements over DeiT-Ti, while effectively bridging the gap between the teacher and student performance for DeiT-S.}
\label{fig:imagenet_plot}
\end{figure}

In this work, we propose a novel projection layer that is derived from a principal concept. Our approach focuses on one key idea: Preserving the intra-batch feature similarity. We highlight that if the similarity between features is preserved, then the projection layer will not change or alter the underlying student representation. This constraint is important since it will maximise the amount of knowledge being distilled to the student backbone. 
For example, if the projector is too expressive it may shortcut the distillation objective by learning some complex non-linear mapping between the two spaces. This result would significantly diminish the efficacy of distillation and is especially detrimental since the projector is thrown away after training.

By enforcing the preservation of the feature similarity, we derive a reparameterisation of the projection layer itself using the set of orthogonal matrices.  
We propose to efficiently implement this reparameterisation by projecting the weights onto $SO(n)$ and then truncating the excess rows. In doing so, we enforce the property of row-wise orthogonality, while avoiding the need to compute any expensive matrix inversions or factorisations. We show that this constraint not only improves the student performance but also improves the training convergence and the efficacy of distilling inductive biases.

Finally, a common component in the application of knowledge distillation for generative tasks is the use of additional auxiliary losses. 
These losses can encourage the generation of diverse features, which will subsequently lead to the generation of more diverse images. 
However, these auxiliary losses often conflict with the distillation objective and will subsequently degrade the student performance.
To address this limitation, we propose a unified framework for incorporating these auxiliary objectives into the distillation loss itself using a task-specific normalisation step.
Thus, we show that simply whitening the teacher features can implicitly encourage feature diversity, while removing the need for fine-tuning the hyperparameters of many additional losses.
We further demonstrate the importance of this whitening step for data-limited image generation, whereby we achieve a consistent and substantial performance improvements over state-of-the-art. 
In summary, our contributions are outlined as follows:

\begin{itemize}
    \item We propose a novel orthogonal projection layer to maximise the knowledge being distilled through to the student backbone.
    \item We complement our projection with a task-wise normalisation that enables knowledge distillation in generative tasks.
    \item We apply our method to a wide range of tasks and modalities, improving over the state-of-the-art by up to $4.4\%$ on ImageNet-1K (see Fig. \ref{fig:imagenet_plot}).
\end{itemize}
\section{Related work}
\label{sec:related_work}

\noindent{\bf Knowledge Distillation}
utilises the knowledge of a pre-trained model as supervision for a much smaller model, which can then enable the application and deployment within resource constrained environments. The field can be broadly divided into two main areas: logits distillation~\cite{Hinton2015DistillingNetwork, cho2019efficacy, Beyer2022KnowledgeConsistent, KimDistillingRelationsv2, niu2022respecting} and feature distillation~\cite{Romero2015FitNets:Nets, Heo2019ADistillation, Tian2019ContrastiveDistillation, Chen2021DistillingReview, yang2021knowledge,yang2022knowledge,Miles2022InformationDistillation}. Logit distillation focuses on classification based tasks and introduces an additional objective to minimise the distance between the student and teacher predictions. This was originally proposed using the KL divergence~\cite{Hinton2015DistillingNetwork}, however it has since been extended using spherical normalisation~\cite{Guo2020ReducingDistillation}, label decoupling~\cite{Zhao2022DecoupledDistillation}, and probability reweighting~\cite{niu2022respecting}. In our work, we focus on feature distillation due to its generality to other tasks~\cite{Chen2021DistillingReview, Miles2023MobileVOS:Distillation} and modalities~\cite{DistilBert2019, Yang2020TextBrewer:Processing}. Unfortunately, there is no underlying metric for the intermediate representation spaces, which has led to many heuristically derived solutions. For example, the hand-crafted FSP matrices were proposed~\cite{Yim2017ALearning} to capture the relation between features before and after a set of residual layers. Similarly, many other works have proposed to transfer knowledge using the construction of various Gram~\cite{Tung2019Similarity-preservingDistillation, instance_relational_kd, He2022FeatureDistillation} or correlation-based matrices~\cite{Li2020LocalCC, Miles2022InformationDistillation, Peng2019CorrelationDistillation}. The activation boundaries have also been shown to be an effective supervisory signal for distillation~\cite{Heo2019KnowledgeNeurons}, along with the gradients to capture the loss landscape~\cite{Zhu2021ComplementaryDistillation, Srinivas2018KnowledgeMatching}. Another line of work can be loosely grouped together by their inspirations from the self-supervision~\cite{Xu2020KnowledgeSelf-supervision, Tian2019ContrastiveDistillation} or information theory literature~\cite{Miles2023MobileVOS:Distillation, Miles2022InformationDistillation}. In contrast to these methods, we take a step back from the conventional curation of hand-crafted relational objects or objectives. We instead derive and demonstrate that an orthogonal projection is much more effective and can be used in conjunction with just a simple $L2$ loss.

\noindent{\bf Self-supervised learning}
describes the family of pretext tasks used to learn good representations of data in the absence of any ground truth labels. This is an important topic with the overwhelming abundance of unlabelled data available and the increasing costs for human annotations. Self-supervised learning shares some significant similarities with the knowledge distillation literature. For example, contrastive learning~\cite{Chen2020ARepresentations, hjelm2019learning, He2020MomentumLearning} has already inspired many distillation methods~\cite{Tian2019ContrastiveDistillation, Chen2020WassersteinDistillation, xu2020knowledge} and asymmetric architectures~\cite{Grill2020BootstrapLearning, Chen2021ExploringLearning, gidaris2021obow, richemond2020byolwobn} are often described as a form of self-distillation~\cite{Bardes2022VICReg:Learning}. However, the most salient overlap with our work instead lies with the use of a predictor. The predictor is a learnable model that maps from the online network to the momentum network. Its usage is very similar to what is described as a projector~\cite{Chen2022ImprovedEnsemble} in the knowledge distillation literature. DirectPred~\cite{Tian2021UnderstandingPairs} explored the training dynamics of this predictor, which allowed the derivation of a closed form solution for its weights. This work was later simplified by either removing the expensive eigen-decomposition~\cite{wang2022demystifying} or using fast matrix iterations~\cite{richemond2023edge}. In contrast to these works, we explore the role of the projector for knowledge distillation. By building upon a simple set of principles for KD, we are able to derive a cheap reparameterisation of the projector weights that can maximise the knowledge transfer.

\begin{figure*}[!thb]
\centering
\resizebox{.9\linewidth}{!}{%
\includegraphics{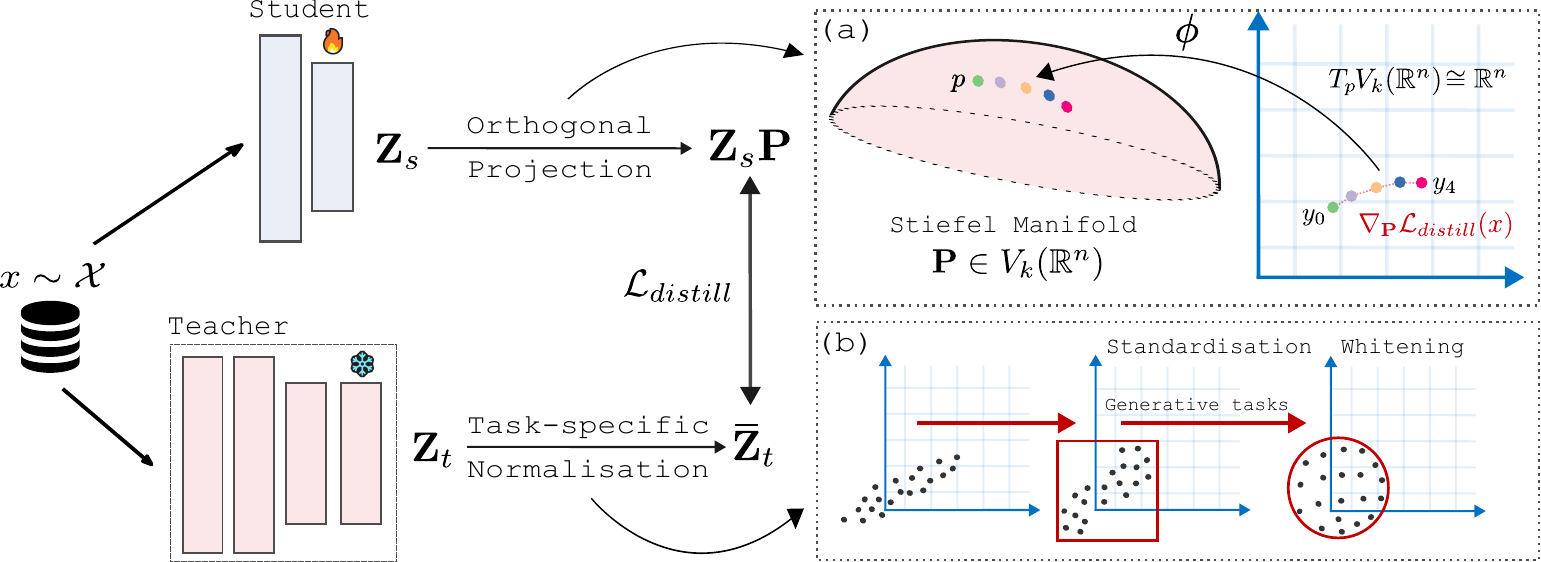}
}
\caption{Illustration of our proposed feature distillation using an orthonormal projection and task-specific feature normalisation. The orthonormal projection (a) maximises the knowledge being distilled to the student backbone, while the task-specific normalisation (b) can introduce domain-specific priors to improve model performance. \fire denotes trainable weights, while \ice denotes weights which are frozen.}
\label{fig:pipeline}
\end{figure*}

\noindent{\bf Layer reparameterisation}
has been widely adopted as a technique for constraining weights to introduce favourable properties. For instance, unitary matrices have been shown to address the gradient issues in RNNs~\cite{arjovsky2016unitary}, positive definite matrices enhance the robustness of batch normalization layers~\cite{reimannian_bn}, and orthogonal matrices offer spectral regularization for improved generalization~\cite{helfrich2018orthogonal, arjovsky2016unitary}. More recent works have shown that low-rank matrices are effective in reducing the cost of fine-tuning large language models~\cite{hu2021lora}, while orthogonal matrices enable the cheap controllable fine-tuning of text-image diffusion models~\cite{qiu2023controlling}. In our work we take these ideas into the context of knowledge distillation with an orthogonal projection. We show that this orthogonal constraint improves both the efficacy of distillation and improves the overall model convergence.
\vspace{-0.4em}
\section{Orthogonal Projections}
\label{sec:method}
\vspace{-0.1em}

Despite the generality of feature distillation, its use is often coupled with many design decisions and heuristics. These decisions arise from the construction of multiple losses between intermediate feature maps which incur significant and unnecessary training overheads. 
To address these constraints, we adopt a simple feature distillation pipeline (see Fig. \ref{fig:pipeline}) using only the features directly before the classifier, or in the case of generative tasks, the latent representation. 

In section \ref{sec:orthonormal} we motivate the necessary conditions to maximise the efficacy of distillation through the projection. This leads to a reparameterisation of the projection as an orthogonal matrix, which is then efficiently implemented in section \ref{sec:orthogonal_reparam}. In section \ref{sec:minimize_redundancy} we provide some interesting additional insights into the properties of these orthogonal projections, while in section \ref{sec:whitening} we extend our distillation pipeline to improve the performance on both generative and discriminative tasks by using an additional task-specific normalisation step.

\subsection{Why use orthogonal projections?} 
\label{sec:orthonormal}

Our main objective is to mitigate the possibility of the projection layer learning any new representation of the data that is not shared by the feature extractor. This is important because the projection layer is thrown away after training and we want to match the feature extractor with the teacher, rather than solely matching the projected features. To achieve this, we propose to preserve the structural information through the projection.
We describe this structural information using a kernel matrix $\mathbf{K} \in \mathbb{R}^{b \times b}$, where $b$ is the batch-size. This kernel matrix captures the pairwise similarity between all features within a batch: 
\vspace{-.5em}

\begin{small}
\begin{align}
    \mathbf{K}_{ij} &= k(\mathbf{Z}^s_i, \mathbf{Z}^s_j) = \braket{\mathbf{Z}^s_i, \mathbf{Z}^s_j}_{\mathcal{H}},
\end{align}
\end{small}

\noindent where $\mathcal{H}$ is some Hilbert space implicitly defined by the positive-definite real-valued kernel $k$ and $\mathbf{Z}^s \in \mathbb{R}^{b \times d_s}$ are the student features of dimension $d_s$. We aim to preserve $\mathbf{K}$ under the application of a linear transformation of its arguments.
This is equivalent to preserving the structural information of the features. We can express many practical kernels, such as the radial basis function kernel or the polynomial kernel, using a Taylor series expansion~\cite{cotter2011explicit}:
\vspace{-.5em}

\begin{small}
\begin{align}
    k(\mathbf{Z}^s_{i}, \mathbf{Z}^s_{j}) &= \sum_{n=0}^{\infty}a_n \braket{\mathbf{Z}^s_{i}, \mathbf{Z}^s_{j}}^n,
\end{align}
\end{small}

\noindent where $a_n$ are the coefficients. This expression shows that we simply need a transformation $\mathbf{P}$ that preserves inner products. Using the canonical inner product in $\mathbb{R}^{d_s}$, we derive this constraint on $\mathbf{P}$ as follows:

\begin{small}
\begin{align}
    \mathbf{Z}^s_i (\mathbf{Z}^s_j)^T &=  \mathbf{Z}^s_i\mathbf{P} (\mathbf{Z}^s_j\mathbf{P})^T \\
    &= \mathbf{Z}^s_i\mathbf{P}\mathbf{P}^T (\mathbf{Z}^s_j)^T,  
\end{align}
\end{small}

\noindent which holds if $\boxed{\mathbf{P}^T = \mathbf{P}^{-1}}$. This constraint conveniently defines the special orthogonal group $SO(d_s)$ with $d_s = d_t$. This group parameterises the set of all rotations in $\mathbb{R}^{d_s}$, thus very naturally and intuitively preserves the idea of structural information. 
However, in the general case where $d_s \neq d_t$, our projection matrix $\mathbf{P}$ is no longer square.
This means that there exists no canonical inverse for our derived constraint to hold. 
Choosing the right-inverse defines the set of matrices with orthonormal rows, whereas choosing the left-inverse defines the set of matrices with orthonormal columns.
Motivated by the need for an efficient reparameterisation, we focus our attention on the right-inverse, which defines the set of matrices with orthonormal rows, its transpose of which is conveniently represented as a Stiefel matrix manifold \cite{edelman1998geometry}, denoted $V_{d_t}(\mathbb{R}^{d_s})$. 
To simplify further notation, we omit this distinction between the two.

Since the Stiefel manifold is smooth, it facilitates the use of standard gradient descent techniques using reparameterisations\footnote{Reparameterisations can be seen as surjective functions that map from Euclidean space back onto the manifold.}. In the next section we provide an efficient implementation of this reparameterisation. 

\subsection{Orthogonal reparameterisation} 
\label{sec:orthogonal_reparam}
There are a few convenient ways to ensure orthogonality of the projection matrix $\mathbf{P}$. One of these ways is to use a Cayley transformation \cite{cayley} that constructs an orthogonal matrix $\mathbf{P}$ from a skew-symmetric matrix: $\mathbf{P} = (\mathbf{I} - \mathbf{W})(\mathbf{I} + \mathbf{W})^{-1}$ where $\mathbf{W} = -\mathbf{W}^T$. Unfortunately, this closed-form parameterisation, despite its simplicity, requires the expensive computation of a large matrix inverse. 
Using a QR decomposition can also be considered, but will require the use of expensive iterative algorithms, such as the Gram Schmidt process.
Since an orthogonal reparameterisations map is
needed for each iteration during training, it is critical for its evaluation to be computationally cheap.
To address this computational constraint we propose an efficient algorithm that avoids the need for any expensive matrix inversions or factorisations.
We propose to instead perform a cheap parameterisation map onto $SO(d_t)$ using the
matrix exponential $exp(\mathbf{A})$, which can be efficiently implemented using the pad\'e approximation. 
Knowing that $\mathbf{W}$ is skew-symmetric, we show that $\exp(\mathbf{W})$ is an orthogonal matrix using a few properties of the exponential:  
$\exp(\mathbf{W}) \cdot \exp(\mathbf{W})^T = \exp(\mathbf{W} + \mathbf{W}^T) = \exp(\mathbf{-W}^T + \mathbf{W}^T) = \exp(\mathbf{0}) = \mathbf{I}$.
We then project back to $V_{d_t}(\mathbb{R}^{d_s})$ by dropping the last $d_t - d_s$ rows. These two sequential steps are given more compactly as follows::

\vspace{-0.7em}
\begin{equation}
    \phi : \mathbf{W} \xrightarrow[]{\exp(\mathbf{W})} \mathbf{A} \in SO(d_t) \xrightarrow[]{\mathbf{A}_{:d_s}} \mathbf{P} \in V_{d_t}(\mathbb{R}^{d_s}),
    \label{eqn:parameterisation_map}
\end{equation}

\noindent where the subscript notation follows from the slicing convention in Pytorch~\cite{paszke2019pytorch} and Numpy~\cite{numpy}. This detour only requires the computation of one exponential, which can be cheaply evaluated using the Pad\'e approximation~\cite{pade_expm}. We show an illustration of this re-parameterisation in Fig. \ref{fig:pipeline}{\color{red} a} with the map $\phi$ described in equation \ref{eqn:parameterisation_map}.

\subsection{Orthogonal projections minimise redundancy}
\label{sec:minimize_redundancy}

\begin{figure}
\centering
\resizebox{.95\linewidth}{!}{%
\includegraphics{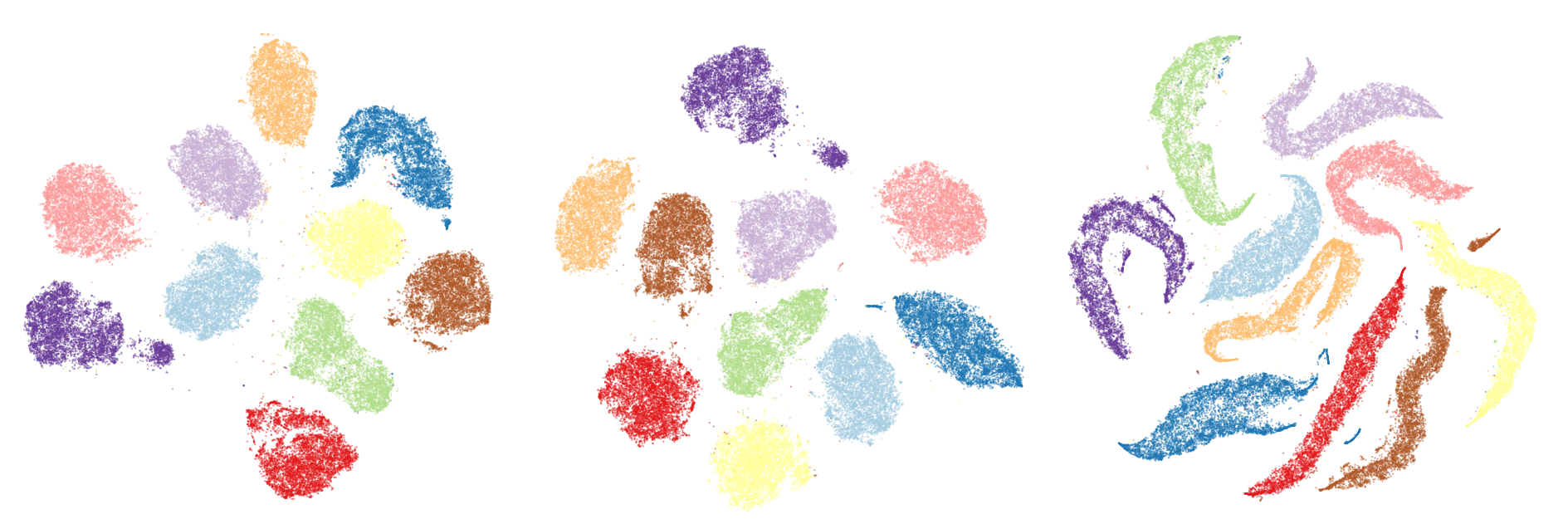}
}
\subfloat[original]{\vspace{-1em}\hspace{.33\linewidth}}
\subfloat[orthogonal]{\vspace{-1em}\hspace{.33\linewidth}}
\subfloat[linear]{\vspace{-1em}\hspace{.33\linewidth}}
\caption{t-SNE visualisation~\cite{tsne} of features undergoing either a linear or orthogonal transformation. The orthogonal transformation preserves all of the structural feature information, whereas the linear projection can distort a lot of structure, which can diminish the efficacy of distillation.}
\label{fig:minimise_redundancy}
\end{figure}

The orthogonal transformations ensure that the projected features $\mathbf{Z}^s\mathbf{P}$ will not only be a linear combination of the original features, but also be a transformation that preserves the notion of distance between features. This result can be explained more concretely by observing that the singular values of $\mathbf{P}$ are all $1$. Geometrically, this means that the projection is not squashing or distorting along any of the dimensions, which would bias the loss toward reconstructing the teachers features using only a subset of features. 
We illustrate this phenomenon in Fig. \ref{fig:minimise_redundancy} where we find that even a simple linear projection distorts the features making them overlap with each other.
This is caused because it attempts to align the space with the teacher. 
This level of distortion can degrade the linear separability of the features for the classifier, which impacts the model performance.
In contrast, the orthogonal projection preserves the underlying feature manifold.

\subsection{Introducing domain-specific priors} 
\label{sec:whitening}
For many tasks it is important to invoke domain specific priors or auxiliary losses to improve model performance~\cite{Liu2019StructuredSegmentation, cui2023kddlgan}. Unfortunately, many of these auxiliary losses conflict with the distillation objective and hinder its efficacy. Instead, we propose a general framework for normalisation that naturally and implicitly incorporates these priors into the distillation objective itself. We show that standardisation is very effective for \underline{discriminative} tasks by improving the model convergence. This convergence property can be attributed to the improved robustness of the distillation loss to random input perturbations. Similarly, we also show that whitening is a critical step for \underline{generative} tasks by providing an implicit and soft encouragement of diverse features, which has been proven effective for generating diverse images~\cite{elfeki2019gdpp}. Whitening can be much more effective and significantly cheaper than introducing additional auxiliary losses, which has been previously proposed in the literature~\cite{cui2023kddlgan}. We now provide a more detailed illustration of these ideas. 

\begin{figure}[H]
\centering
\resizebox{.9\linewidth}{!}{%
\includegraphics{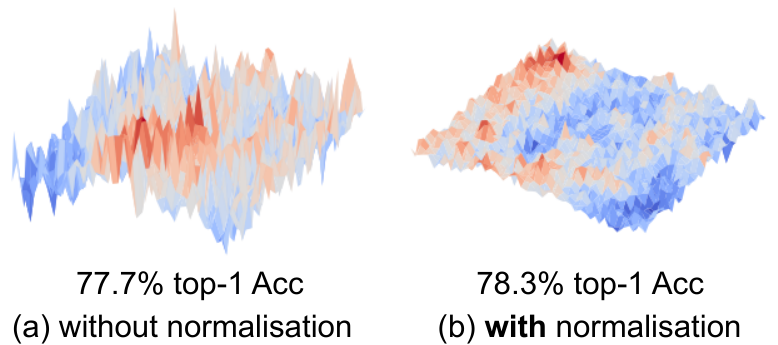}
}
\caption{Visualisation of the $V_k$D-Ti $\mathcal{L}_{distill}$ loss landscape with perturbations of the input image across two random dimensions. Normalisation significantly reduces the sensitivity of the loss to random perturbations, which leads to improved robustness and convergence for training.}
\label{fig:faster_convergence}
\end{figure}

\vspace{-1.8em}
\paragraph{Standardisation improves model convergence.}
In our application of knowledge distillation to discriminative tasks, we observe that a straightforward normalisation of the teacher's representation yields a notable improvement in the robustness of the distillation loss to spurious deformations of the input image (see Fig. \ref{fig:faster_convergence}). These spurious deformations arise from the increasingly expressive families of data augmentation strategies being commonly employed for knowledge distillation~\cite{Touvron2021TrainingAttention}. We find that minimizing this loss variance can significantly improve the overall model convergence and performance.

\vspace{-1em}
\paragraph{Whitening improves feature diversity.}
By whitening the teacher features, we derive a lower bound that resembles a feature diversity loss~\cite{cui2023kddlgan}. We start with an $L2$ loss between $\mathbf{Z}_s\mathbf{P} \in \mathbb{R}^{b \times d}$ and $\mathbf{Z}_t \in \mathbb{R}^{b \times d}$. Since $\mathbf{P}$ is an orthogonal projection that preserves inner products (see section \ref{sec:orthonormal}), we omit its usage to simplify analysis:

\begin{small}
\begin{align}
    \mathcal{L}_{distill} &= \lpnorm{\mathbf{Z}^s - \mathbf{Z}^t}^2 \nonumber \\
    &= \sum_{i \neq j}\lpnorm{\mathbf{Z}^s_{:,j} - \mathbf{Z}^t_{:,i} - \mathbf{Z}^t_{:,j} + \mathbf{Z}^t_{:,i}}^2 \nonumber \\
    &= \sum_{i \neq j}\lpnorm{\mathbf{Z}^s_{:,j} - \mathbf{Z}^t_{:,i}}^2 + \lpnorm{\mathbf{Z}^t_{:,j} + \mathbf{Z}^t_{:,i}}^2 \\
    & \;\;\;\;\;\;\;\;\;\;\;\;\;\;\;\;\;\;\;\;\;\;\;\;\;\;\;\;\;\;\;\; -2\braket{\mathbf{Z}^s_{:,j} - \mathbf{Z}^t_{:,i}, \mathbf{Z}^t_{:,j} + \mathbf{Z}^t_{:,i}} \nonumber
\end{align}
\end{small}

\noindent Since $\mathbf{Z}^t$ is whitened such that $(\mathbf{Z}^t)^T(\mathbf{Z}^t) = \mathbf{I}$, i.e., perfect decorrelation of features, we can significantly simplify the expression above:

\begin{small}
\begin{align}
   &\mathcal{L}_{distill} = \sum_{i \neq j}\lpnorm{\mathbf{Z}^s_{:,j} - \mathbf{Z}^t_{:,i}}^2 + 2 - 2\braket{\mathbf{Z}^s_{:,j} - \mathbf{Z}^t_{:,i}, \mathbf{Z}^t_{:,j} + \mathbf{Z}^t_{:,i}} \nonumber \\
    &\geq \sum_{i \neq j}\lpnorm{\mathbf{Z}^s_{:,j} - \mathbf{Z}^t_{:,i}}^2 + 2- 2\lpnorm{\mathbf{Z}^s_{:,j} - \mathbf{Z}^t_{:,i}}^2\lpnorm{\mathbf{Z}^t_{:,j} + \mathbf{Z}^t_{:,i}}^2 \nonumber  \\
    &= \boxed{\text{const} - \lambda \sum_{i \neq j} \mathbf{C}_{j, i}^2}, \text{\; where \; } \mathbf{C}_{i, j} = \lpnorm{\mathbf{Z}_{:,j}^s - \mathbf{Z}_{:,i}^t}
\end{align}
\end{small} 

where \textit{const} and $\lambda \geq 3$ are both constants that do not depend on the model parameters. Here $\mathbf{C}$ is the euclidean cross-correlation matrix that captures the distance between all the pairs of student and teacher features.
This derivation simply shows that minimising an $L2$ loss subject to an explicit whitening constraint on the teacher features provides a cross-feature objective.
This cross-feature objective maximises the off-diagonal entries in the cross-correlation matrix, thus encouraging all the features to be decorrelated with respect to the teacher. We describe this process of encouraging decorrelation of features as increasing the feature diversity. 
We validate this connection of whitening with feature diversity in section \ref{sec:data_limited_image_generation}, where we employ knowledge distillation in the context of data-efficient image generation.
\vspace{-1.5em}
\section{Experiments}
\label{sec:experiments}

In this section, we evaluate the generality of our simple knowledge distillation pipeline across three distinct vision tasks: Image classification, object detection, and image generation. In each of these tasks, we consider the harder distillation settings, such as distilling cross-architecture, or in the data-efficient regimes. Throughout we use \colorbox{myblue!100}{$V_k$D} to denote our KD method using an orthogonal projection, i.e., a matrix projection from the Stiefel manifold $\mathbf{V_k(\mathbb{R}^d)}$.

\begin{table}[H]
    \vspace{1em}
    \centering
    \resizebox{1.\linewidth}{!}{%
    \begin{tabular}{lccc}
    
    \toprule
    \belowrulesepcolor{mygrey} 
    \rowcolor{mygrey} Network & acc@1 & Teacher & \#params \\
    \aboverulesepcolor{mygrey}
    \midrule
    RegNetY-160~\cite{Radosavovic2020DesigningSpaces} \textit{\footnotesize CVPR20} & 82.6 & none & 84M \\
    CaiT-S24~\cite{touvron2021going} \textit{\footnotesize ICCV21} & 83.4 & none & 47M \\
    DeiT3-B~\cite{touvron2022deit} \textit{\footnotesize ECCV22} & \textbf{83.8} & none & 87M \\
    \midrule
    DeiT-Ti~\cite{Touvron2021TrainingAttention} \textit{\footnotesize ICML21} & 72.2 & none & 5M\\
    CivT-Ti~\cite{Ren2022Co-advise:Distillation} \textit{\footnotesize CVPR22} & 74.9 & regnety-600m + rednet-26 & 6M \\
    MaskedKD~\cite{son2023maskedkd} \textit{\footnotesize arXiv23} & 75.4 & cait-s24 & 6M \\
    Manifold~\cite{hao2022learning} \textit{\footnotesize NeurIPS22} & 76.5 & cait-s24 & 6M \\
    DeiT-Ti\alambicdeit\xspace ~\cite{Touvron2021TrainingAttention} \textit{\footnotesize ICML21} & 74.5 & regnety-160 & 6M \\
    \myalign{l}{\;\;\;\footnotesize \rotatebox[origin=c]{180}{$\Lsh$} 1000 epochs } & 76.6 & regnety-160 & 6M \\
    DearKD~\cite{chen2022dearkd} \textit{\footnotesize CVPR22} & 74.8 & regnety-160 & 6M \\
    \myalign{l}{\;\;\;\footnotesize \rotatebox[origin=c]{180}{$\Lsh$} 1000 epochs } & 77.0 & regnety-160 & 6M \\
    USKD~\cite{yang2023knowledge} \textit{\footnotesize ICCV23} & 75.0 & regnety-160 & 6M \\
    SRD~\cite{miles2024understanding} \textit{\footnotesize AAAI24} & 77.2 & regnety-160 & 6M \\
    \rowcolor{myblue} $V_k$D-Ti & \textbf{78.3} & regnety-160 & 6M \\
    \aboverulesepcolor{myblue}
    \midrule
    DeiT-S~\cite{Touvron2021TrainingAttention} \textit{\footnotesize ICML21} & 79.8 & none & 22M\\
    CivT-S~\cite{Ren2022Co-advise:Distillation} \textit{\footnotesize CVPR22} & 82.0 & regnety-4gf + rednet-50 & 22M \\
    MaskedKD~\cite{son2023maskedkd} \textit{\footnotesize arXiv23} & 81.4 & deit3-b & 22M \\
    DeiT-S\alambicdeit\xspace ~\cite{Touvron2021TrainingAttention} \textit{\footnotesize ICML21} & 81.2 & regnety-160 & 22M \\
    \myalign{l}{\;\;\;\footnotesize \rotatebox[origin=c]{180}{$\Lsh$} 1000 epochs } & 82.6 & regnety-160 & 22M \\
    DearKD~\cite{chen2022dearkd} \textit{\footnotesize CVPR22} & 81.5 & regnety-160 & 22M \\
    \myalign{l}{\;\;\;\footnotesize \rotatebox[origin=c]{180}{$\Lsh$} 1000 epochs } & 82.8 & regnety-160 & 22M \\  
    USKD~\cite{yang2023knowledge} \textit{\footnotesize ICCV23} & 80.8 & regnety-160 & 22M \\
    SRD~\cite{miles2024understanding} \textit{\footnotesize AAAI24} & 82.1 & regnety-160 & 22M \\
    \rowcolor{myblue} $V_k$D-S & \textbf{82.3} & regnety-160 & 22M \\
    \aboverulesepcolor{myblue}
    \bottomrule
\end{tabular}
    }
    \vspace{-0.5em}
    \caption{Data-efficient training of transformers using knowledge distillation on the ImageNet-1K dataset. Unless specified, each model is only trained for 300 epochs.}
    \label{table:deit_simple}
\end{table}

\begin{table*}[!t]
    \centering
    \footnotesize
    \begin{tabular}{lllllllllll}
    \toprule
    \belowrulesepcolor{mygrey} 
    \rowcolor{mygrey} Method & Backbone & Epochs & AP & AP$_{50}$ & AP$_{75}$ & AP$_{S}$ & AP$_{M}$ & AP$_{L}$ & Param. & FPS \\
    \aboverulesepcolor{mygrey}
    \midrule
    & Swin-nano & 50 & 40.4 & 59.6 & 43.3 & 23.2 & 42.5 & 55.8 & 16M & 20.0 (45.8) \\
    \rowcolor{myblue} \cellcolor{white} & \myalign{l}{\;\;\;\footnotesize \rotatebox[origin=c]{180}{$\Lsh$} w/ $V_k$D } & 50 & \textbf{43.0} & \textbf{62.3} & \textbf{46.2} & \textbf{24.8} & \textbf{45.3} & \textbf{60.1} & & \\
    & Swin-tiny & 50 & 44.8 & 64.5 & 48.7 & 25.9 & 47.6 & 62.1 & 38M & 17.2 (26.5) \\
    \rowcolor{myblue} \cellcolor{white} & \myalign{l}{\;\;\;\footnotesize \rotatebox[origin=c]{180}{$\Lsh$} w/ $V_k$D } & 50 & \textbf{46.9} & \textbf{66.6} & \textbf{50.9} & \textbf{27.8} & \textbf{49.8} & \textbf{64.6} & & \\
    & Swin-small & 50 & 47.5 & 67.7 & 51.4 & 29.2 & 50.7 & 64.8 & 61M & 12.1 (16.5) \\ 
    \rowcolor{myblue} \multirow{-6}*{ViDT} \cellcolor{white} & \myalign{l}{\;\;\;\footnotesize \rotatebox[origin=c]{180}{$\Lsh$} w/ $V_k$D } & 50 & \textbf{48.5} & \textbf{68.4} & \textbf{52.4} & \textbf{30.8} & \textbf{52.2} & \textbf{66.0} & & \\
    \bottomrule   
\end{tabular}
    \caption{Comparison with other detectors on COCO2017 val set. FPS is measured with batch size 1 of 800 × 1333 resolution on a single Tesla V100 GPU, where the value inside the parentheses is measured with batch size 4 of the same resolution to maximize GPU utilisation. All of the student models are distilled from a pre-trained ViDT-base.}
    \label{table:vidt}
    \vspace{0.5em}
\end{table*}

\begin{table}[!t]
    \centering
    \resizebox{1.\linewidth}{!}{%
    \begin{tabular}{lcccc}
    \toprule
    \belowrulesepcolor{mygrey} 
    \rowcolor{mygrey} Student & \multicolumn{2}{c}{ViDT (Swin-nano)} & \multicolumn{2}{c}{ViDT (Swin-tiny)}  \\
    \aboverulesepcolor{mygrey}
    \midrule
    Teacher & \begin{tabular}{@{}c@{}}ViDT \\ (small)\end{tabular} & \begin{tabular}{@{}c@{}}ViDT \\ (base)\end{tabular} & \begin{tabular}{@{}c@{}}ViDT \\ (small)\end{tabular} & \begin{tabular}{@{}c@{}}ViDT \\ (base)\end{tabular} \\
    \midrule
    No Distillation~\cite{song2021vidt} \textit{\footnotesize ICLR22} & \multicolumn{2}{c}{40.4} & \multicolumn{2}{c}{44.8} \\
    Token Matching~\cite{song2021vidt} \textit{\footnotesize ICLR22} & 41.5 & 41.9 & 45.8 & 46.5 \\
    \rowcolor{myblue} $V_k$D & \textbf{42.2} & \textbf{43.0} & \textbf{45.9} & \textbf{46.9} \\
    \aboverulesepcolor{myblue}
    \bottomrule
\end{tabular}
    }
    \caption{Comparison of ViDT on COCO2017 val set. We report AP for the student models distilled from different teacher models.}
    \label{table:vidt_simple}
\end{table}

\vspace{-2em}
\paragraph{Implementation details.}
We train all models in Pytorch~\cite{Paszke2017AutomaticPyTorchv2} using 2 NVIDIA V100 GPUs. 
For the ImageNet experiments, we follow DeiT~\cite{Touvron2021TrainingAttention} using the same training schedule and optimization parameters. We also adopt Mixup~\cite{Zhang2017Mixup:Minimization} augmentation, but replace rand-augment with random gray scaling, gaussian blurring, and solarization.
We use AdamW~\cite{loshchilov2017decoupled} optimizer with learning rate set to $0.001$ and weight decay to $0.05$.
For the object detection experiments, we follow the same training methodology as ViDT~\cite{song2021vidt} except that we replace the original token matching loss with our $V_kD$. Finally, for the image generation task, we use the same training methodology as KD-DLGAN~\cite{cui2023kddlgan} except that we remove the auxiliary diversity losses and instead replace it with either teacher standardisation or whitening. We also remove any distillation from the text encoders, thus further reducing the cost of our method in comparison.

\subsection{Data efficient training of transformers}
\label{sec:data_efficient_transformers}
We experiment with vision transformers, due to their proven success across a variety of fields. However, despite this success, they demand excessive training data and long training schedules. This limitation has motivated the use of knowledge distillation for improving the data efficiency of transformer models~\cite{Touvron2021TrainingAttention}.
We compare our method with several others using the common knowledge distillation setting proposed alongside DeiT ~\cite{Touvron2021TrainingAttention} that uses a CNN teacher pre-trained on ImageNet-21K.
We train each student model for 300 epochs on ImageNet-1K for the image classification task.
Unlike other methods that propose to leverage the efficacy of distilling through distillation tokens alone, we propose to distill directly through to the patch tokens.
We present the results in Tab. \ref{table:deit_simple} where we massively outperform the previous state-of-the-art.
In the tiny architecture, we outperform the baseline by $6.1$ percentage points ($pp$), and the previous best method that uses the same teacher, the USKD by $3.3pp$, or a relative improvement of $4.4\%$.
Interestingly, we perform better than DearKD trained for $1000$ epochs by $1.3pp$, showing that for distillation, it is not necessary to train that long.
We reach similar results in the small architecture too, where we outperform all the other methods that use similar training resources, and reach competitive results with the methods that use more than $3$ times as long training time.
In fact, unlike other methods, our approach bridges the gap between the teacher model that reaches $82.6\%$ accuracy without needing to introduce any excessively long training schedules.

\begin{table*}[!t]
    \centering
    \footnotesize
    \begin{tabular}{lcccccc}
    \toprule
    \belowrulesepcolor{mygrey}
    \rowcolor{mygrey} & & CIFAR-10 & & & CIFAR-100 & \\ 
    \rowcolor{mygrey} \multirow{-2}{*}{Method} & 10\% Data & 20\% Data & 100\% Data & 10\% Data & 20\% Data & 100\% Data \\
    \aboverulesepcolor{mygrey}
    \midrule
    DA + KD (CLIP) & $22.03 \pm 0.07$ & $13.70 \pm 0.08$ &  $8.70 \pm 0.02$ & $33.93 \pm 0.09$ & $21.76 \pm 0.06$ & $11.74 \pm 0.02$ \\
    
    DA (Baseline) & $23.34 \pm 0.09$ & $14.53 \pm 0.10$ & $8.75 \pm 0.03$ & $35.39 \pm 0.08$ & $22.55 \pm 0.06$ & $11.99 \pm 0.02$ \\
    
    KD-DLGAN & $14.20 \pm 0.06$ & $11.01 \pm 0.07$ & $8.42 \pm 0.01$ & $18.03 \pm 0.11$ & $15.60 \pm 0.08$ & $\mathbf{10.28 \pm 0.03}$ \\

    \rowcolor{myblue} DA + $V_k$D & $16.42 \pm 0.07$ & $10.94 \pm 0.07$ & $8.28 \pm 0.01$ & $24.92 \pm 0.15$ & $17.97 \pm 0.17$ & $10.61 \pm 0.08$ \\
    
    \rowcolor{myblue} \myalign{l}{\;\;\;\footnotesize \rotatebox[origin=c]{180}{$\Lsh$} w/ whitening} & $\mathbf{13.16 \pm 0.06}$ & $\mathbf{10.24 \pm 0.06}$ & $\mathbf{8.20 \pm 0.03}$ & $\mathbf{16.87 \pm 0.09}$ & $\mathbf{14.00 \pm 0.07}$ & $10.41 \pm 0.01$ \\
    
    \aboverulesepcolor{myblue}
    \bottomrule
\end{tabular}
    \vspace{-0.5em}
    \caption{Comparison with the state-of-the-art over CIFAR-10 and CIFAR 100. Competitive performance is achieved using the orthogonal projection alone, however, introducing a simple whitening step is sufficient in outperforming state-of-the-art by a significant margin. All the compared methods employ BigGAN as the backbone. FID is averaged over three runs.}
    \vspace{0.5em}
    \label{table:kdgan}
\end{table*}

\subsection{Object detection}
\label{sec:object_detection_and_instance_segmentation}
We consider the object detection task using the common MS-COCO benchmark ~\cite{lin2015microsoft}.
We use the ViDT transformer architecture~\cite{song2021vidt} due to its task performance and its efficiency on consumer hardware.
We present the results in Tab. \ref{table:vidt}, and observe a significant and consistent improvement across a wide range of different ViDT variants.
We improve using Swin-nano backbone by $2.6pp$, in Swin-tiny backbone by $2.1pp$ and in Swin-small backbone by $1pp$.
Furthermore, we also compare against an alternative distillation method, described as token matching \cite{song2021vidt}.
We present these results in Tab. \ref{table:vidt_simple}.
Our method outperforms token matching by up to $1.1pp$, reaching the best results when we use a larger teacher, demonstrating that our method is not limited in the cases of larger capacity gaps.

\subsection{Data limited image generation}
\label{sec:data_limited_image_generation}
To demonstrate the generality of our feature distillation framework, we consider an image generation task and compare to the recent KD-DLGAN~\cite{cui2023kddlgan}.
KD-DLGAN proposes to use both the text and feature embeddings for the feature guidance followed by an additional diversity loss.
Using our novel framework, we show that neither of these explicit additional losses is necessary. Instead, we use a simple whitening of features to encourage the generation of diverse images.
We show these results in Tab. \ref{table:kdgan}. A noticeable observation in these results is that whitening can obtain the most significant improvements in the more extreme data-limited regimes. For example, when training on 10\% data with CIFAR-100, whitening the teacher features can improve the FID by up to 9.09. 
This result highlights that feature diversity is much more critical when there is insufficient training data.
To show that our method is much more general than other KD methods in the literature, we also include a comparison for the hardest data-efficient regime in Tab. \ref{table:kdgan_small}. The results show a consistent improvement in performance, which highlight the importance of engineering both the projector architecture and the normalisation scheme, as opposed to focusing on the distance metrics alone~\cite{Tung2019Similarity-preservingDistillation, cui2023kddlgan}.

\begin{table}
    \centering
    \footnotesize
    \begin{tabular}{lcc}
    \toprule
    \belowrulesepcolor{mygrey}
    \rowcolor{mygrey} & CIFAR-10 & CIFAR-100 \\ 
    \rowcolor{mygrey} \multirow{-2}{*}{Method} & 10\% Data & 10\% Data \\
    \aboverulesepcolor{mygrey}
    \midrule
    DA (Baseline)~\cite{zhao2020differentiable} \textit{\footnotesize NeurIPS20} & $23.34 \pm 0.09$ & $35.39 \pm 0.08$ \\
    FitNets~\cite{Romero2015FitNets:Nets} \textit{\footnotesize ICLR14} & $22.03 \pm 0.07$ & $33.93 \pm 0.09$ \\
    Label Distillation~\cite{Hinton2015DistillingNetwork} \textit{\footnotesize arXiv15} & $20.46 \pm 0.10$ & $34.14 \pm 0.11$ \\
    PKD~\cite{Passalis2018LearningTransfer} \textit{\footnotesize ECCV19} & $21.34 \pm 0.08$ & $32.15 \pm 0.13$ \\
    SPKD~\cite{Tung2019Similarity-preservingDistillation} \textit{\footnotesize ICCV19} & $19.11 \pm 0.07$ & $31.97 \pm 0.10$ \\
    KD-DLGAN~\cite{cui2023kddlgan} \textit{\footnotesize CVPR23} & $14.20 \pm 0.06$ & $18.03 \pm 0.11$ \\

    \rowcolor{myblue} $V_k$D & $16.47 \pm 0.07$ & $24.92 \pm 0.15$ \\

    \rowcolor{myblue} \myalign{l}{\;\;\;\footnotesize \rotatebox[origin=c]{180}{$\Lsh$} w/ whitening} & $\mathbf{13.16 \pm 0.06}$ & $\mathbf{16.87 \pm 0.09}$ \\
    \aboverulesepcolor{myblue}
    \bottomrule
\end{tabular}
    \caption{Comparison to other knowledge distillation methods for image generation. Results were originally reported in \cite{cui2023kddlgan} and highlight the importance of incorporating domain-specific priors - in this case, encouraging diverse features.}
    \label{table:kdgan_small}
\end{table}

\subsection{Ablation study}
\label{sec:ablation_experiments}
We do a series of ablation studies to highlight the importance of our proposed building blocks. We also provide qualitative results that provide additional insights into explaining the efficacy of our distillation framework.

\begin{figure}
    \centering
    \resizebox{1.\linewidth}{!}{%
    \includegraphics{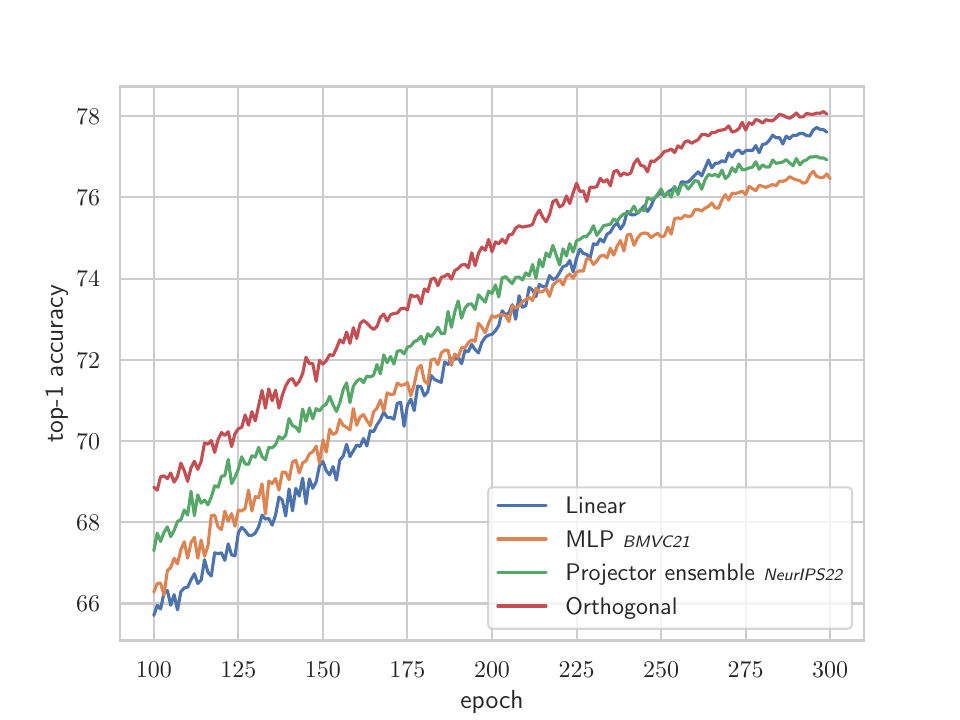}
    }
    \caption{Comparing the performance and convergence of various projector reparameterisations. Although the MLP layer initially trains fast, it begins to saturate as it starts to learn a new representation of the data.}
    \label{fig:projector_ablation}
\end{figure}

\vspace{-0.8em}
\paragraph{Effectiveness of orthogonal projections.} 
\label{sec:projection_ablation}
To demonstrate the effectiveness of constraining the projection weights to be orthogonal, we consider the use of various other projection variants. We analyze the use of a projector ensemble~\cite{Chen2022ImprovedEnsemble}, a multi-layer perceptron~\cite{Navaneet2021SimReg:Distillation}, and a standard linear layer. 
We present the results in Fig. \ref{fig:projector_ablation}.
We observe that both the MLP and projector ensembles show improved performance over a linear layer when under short training schedules.
However, when we extend the training schedule, the linear layer becomes much more effective.
This is a consequence of the expressive projections beginning to learn new representations that are no longer shared by the student feature extractor.
In contrast, our orthogonal projection not only improves the final accuracy but also improves the convergence properties for training, reaching state-of-the-art results in only $\sim 200$ epochs.

\paragraph{The effect of each block. }
We now disentangle the contribution of each block of our framework.
In table \ref{table:ablation:metric} we report the final accuracy with and without normalisation or an orthogonal projection. 
We observe that, in the classification task, most of the performance improvement in obtained by our orthogonal projection.
For example, the orthogonal projection alone boosts the performance from $76.3\%$ to $77.9\%$.
This observation is in contrast to the generative tasks, whereby we observe a necessity to use normalisation for strong performance.

\begin{table}[H]
    \centering
    \footnotesize
    \begin{tabular}{cccc}
    \toprule
    \belowrulesepcolor{mygrey} 
    \rowcolor{mygrey} Orth. & Norm. & Acc@1 & Acc@5 \\
    \aboverulesepcolor{mygrey}
    \midrule
    {\color{crossred} \xmark} & {\color{crossred} \xmark} & 76.3 & 93.3 \\  %
    {\color{crossred} \xmark} & {\color{tickgreen} \cmark} & 76.9 & 93.5 \\  
    {\color{tickgreen} \cmark}  & {\color{crossred} \xmark} & 77.9 & 93.9 \\  
    \rowcolor{myblue} {\color{tickgreen} \cmark} & {\color{tickgreen} \cmark} & \textbf{78.3} & \textbf{94.1} \\  
    \aboverulesepcolor{myblue}
    \bottomrule
\end{tabular}
    \caption{Highlighting the primary importance of an orthonormal projection. Image classification on ImageNet-1K using a DeiT-Ti student and a RegNety-160 teacher.}
    \label{table:ablation:metric}
    \vspace{-0.5em}
\end{table}

\vspace{-0.8em}
\paragraph{Whitening for generative tasks.} To empirically confirm the importance of feature whitening for generative tasks, we perform an evaluation with and with it being used. We present the results in Tab. \ref{table:kdgan} and \ref{table:kdgan_small}. 
We observe that not only is whitening necessary to outperform previous state-of-the-art image generation, but it also leads to a larger increase in performance in the data-limited regime. This result highlights the more prominent importance of diverse features when limited training data is available.

\begin{figure}
    \centering

    \captionsetup[subfigure]{labelformat=empty}
    \centering
    \subfloat[Input]{\hspace{.33\linewidth}}\hfill
    \subfloat[Deit-Ti\alambicdeitcol\xspace]{\hspace{.33\linewidth}}\hfill
    \subfloat[$V_k$D-Ti]{\hspace{.33\linewidth}}
    \\  
    
    \includegraphics[width=.33\linewidth]{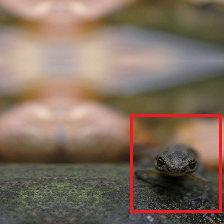}\hfill
    \includegraphics[width=.33\linewidth]{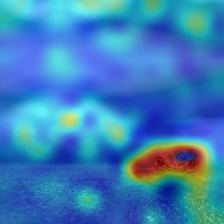}\hfill
    \includegraphics[width=.33\linewidth]{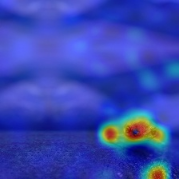}\hfill
    \includegraphics[width=.33\linewidth]{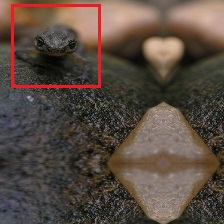}\hfill
    \includegraphics[width=.33\linewidth]{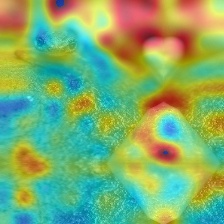}\hfill
    \includegraphics[width=.33\linewidth]{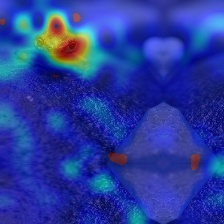}\hfill
    \vspace{-0.5em}
    \caption{Evaluating the translational equivariance of attention maps. We select the best channel for the first translation and observe its attention maps after translating the input image again.}
    \vspace{0.5em}
    \label{fig:translation_ablation}
\end{figure}

\vspace{-0.5em}
\paragraph{Distilling inductive biases.} 
\label{sec:inductive_biases}
Knowledge distillation has proven effective for improving the data efficiency in training transformer models, especially when the teacher is a CNN. 
Unfortunately, there has been little qualitative analysis on explaining why this cross-architecture setting helps. 
We now quantify that this result is a consequence of providing a soft distillation of inductive biases (in this case translational equivariance).
In Fig. \ref{fig:translation_ablation}, we explore the impact of applying a translation on the attention maps of a given layer. 
We observe that any translation of an object is reflected with a translation of the attention maps.
Interestingly, this is unlike other methods, such as Deit-Ti\alambicdeit\xspace, where the attention maps become messy after the translation of the original image. This observation suggests that their improvements may instead be attributed to some other factor, such as an implicit regularisation of the model.

\begin{figure}[!t]
    \centering

    \captionsetup[subfigure]{labelformat=empty}
    \centering
    \subfloat[Input]{\hspace{.25\linewidth}}\hfill
    \subfloat[Deit-Ti\alambicdeitcol\xspace]{\hspace{.25\linewidth}}\hfill
    \subfloat[CiT-S] {\hspace{.25\linewidth}}\hfill
    \subfloat[$V_k$D-Ti]{\hspace{.25\linewidth}}
    \\   
    \includegraphics[width=.25\linewidth]{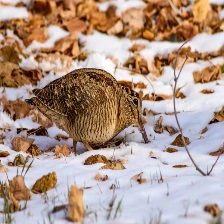}\hfill
    \includegraphics[width=.25\linewidth]{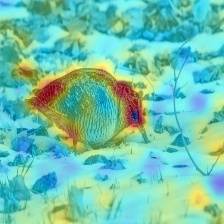}\hfill
    \includegraphics[width=.25\linewidth]{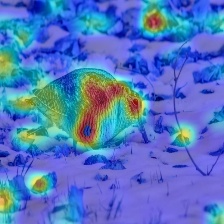}\hfill
    \includegraphics[width=.25\linewidth]{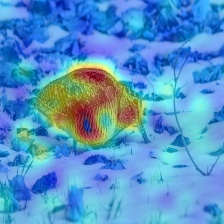}
    \\
    \includegraphics[width=.25\linewidth]{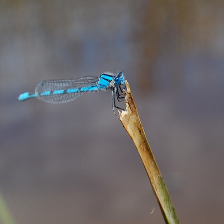}\hfill
    \includegraphics[width=.25\linewidth]{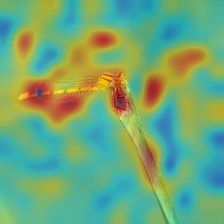}\hfill
    \includegraphics[width=.25\linewidth]{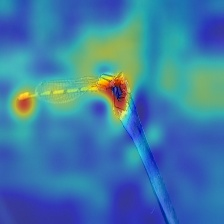}\hfill
    \includegraphics[width=.25\linewidth]{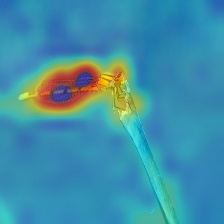}\hfill
    \\
    \includegraphics[width=.25\linewidth]{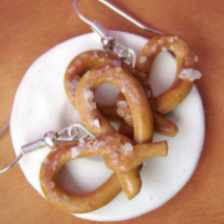}\hfill
    \includegraphics[width=.25\linewidth]{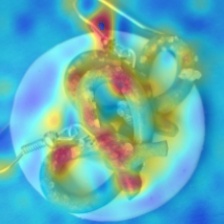}\hfill
    \includegraphics[width=.25\linewidth]{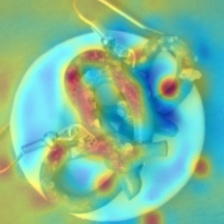}\hfill
    \includegraphics[width=.25\linewidth]{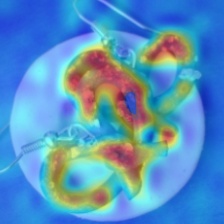}\hfill
    \\
    \includegraphics[width=.25\linewidth]{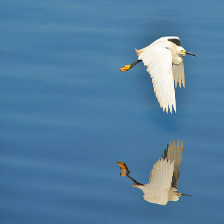}\hfill
    \includegraphics[width=.25\linewidth]{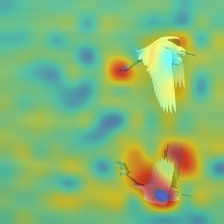}\hfill
    \includegraphics[width=.25\linewidth]{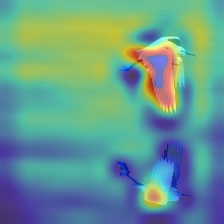}\hfill
    \includegraphics[width=.25\linewidth]{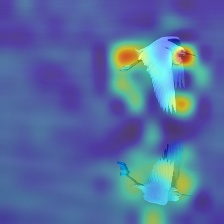}\hfill
    \vspace{-0.5em}
    \caption{Qualitative comparison to other transformer distillation methods. The best channel is selected qualitatively for all examples shown. $V_k$D-Ti is compared against both the same size Deit-Ti and a much larger (3.7$\times$) CiT-S. Best viewed in colour.}
    \vspace{0.5em}
    \label{fig:qualitative_attention}
\end{figure}

\vspace{-0.7em}
\paragraph{Improved localisation of attention maps.} 
\label{sec:attention_scores}
We provide further insights into our orthogonal feature distillation method by analyzing the attention maps of various images. These attention maps show how well the model is attending to the salient objects in an image~\cite{caron2021emerging}.
We compare with various other distillation methods and show the results in Fig. \ref{fig:qualitative_attention}.
We observe that the attention maps of our method are clustered around the boundaries of the objects, unlike the other two methods where the attention maps are spread over the entire image.
In fact, we observe that our distilled model can attend much more to the salient object than the much larger CiT-S model.

\vspace{-0.7em}
\paragraph{Architecture agnostic.}
Our method is agnostic in the choice of features and can be applied to various classifiers, object detectors, or generative models.
For example, in section \ref{sec:data_efficient_transformers}, we distill from a CNN to a transformer, in section \ref{sec:object_detection_and_instance_segmentation}, we distill between two transformers, and in section \ref{sec:data_limited_image_generation} we perform distillation in the other direction, from a transformer to a CNN.
\section{Conclusion}
\label{sec:conclusion}
In this work, we present a novel projection layer with a principled foundation centered on preserving the intra-batch feature similarity. The core idea of maintaining feature similarity ensures that the projection layer does not distort the underlying student representation, thus maximizing the knowledge transfer to the student backbone. We show that enforcing this constraint is equivalent to parameterising the projection weights to have orthonormal rows or columns. Our simple drop-in replacement for the projection layer leads to improved performance across a wide range of distillation tasks, from image classification to object detection. To further improve the generality of this framework, we show that whitening the teachers' features is sufficient and more effective in extending to generative tasks than other methods. 
We show in the experiments that our method improves state-of-the-art by up to $4.4\%$ for image classification and $2.6\%$ for object detection.

{\small
\bibliographystyle{ieee_fullname}
\bibliography{reference,refs}
}

\newpage

\section{Supplementary Material}
We provide more details on the datasets, architectures, and training pipelines. We also provide results on the much smaller CIFAR100 distillation benchmark, where we show competitive or improved performance over state-of-the-art. Finally, we include some qualitative results for the image generation and the complete derivations for orthogonality and whitening.

\subsection{Datasets}
We conduct experiments over a few widely adopted datasets including CIFAR~\cite{Krizhevsky2009LearningImages}, ImageNet~\cite{Russakovsky2014ImageNetChallenge}, and COCO~\cite{Lin2014MicrosoftContext}.

\vspace{-1em}
\paragraph{CIFAR} classification consists of 60K 32×32 RGB images across either 10 or 100 classes with a 5:1 training/testing split. The models are each trained with 100\%, 20\% or 10\% training images~\cite{cui2023kddlgan}.

\vspace{-1em}
\paragraph{ImageNet} classification uses 1.3 million images from 1000 different classes. In these experiment, we set the input size to $224 \times 224$, and follow the same training pipeline and augmentations provided by DeiT~\cite{touvron2022deit}.

\vspace{-1em}
\paragraph{COCO} includes a large-scale object detection benchmark, which we use to evaluate the ViDT model variants. It consists of 330k images with 80 different object categories.

\subsection{Handcrafted projections}
\begin{figure}
\centering
\resizebox{.99\linewidth}{!}{%
\includegraphics{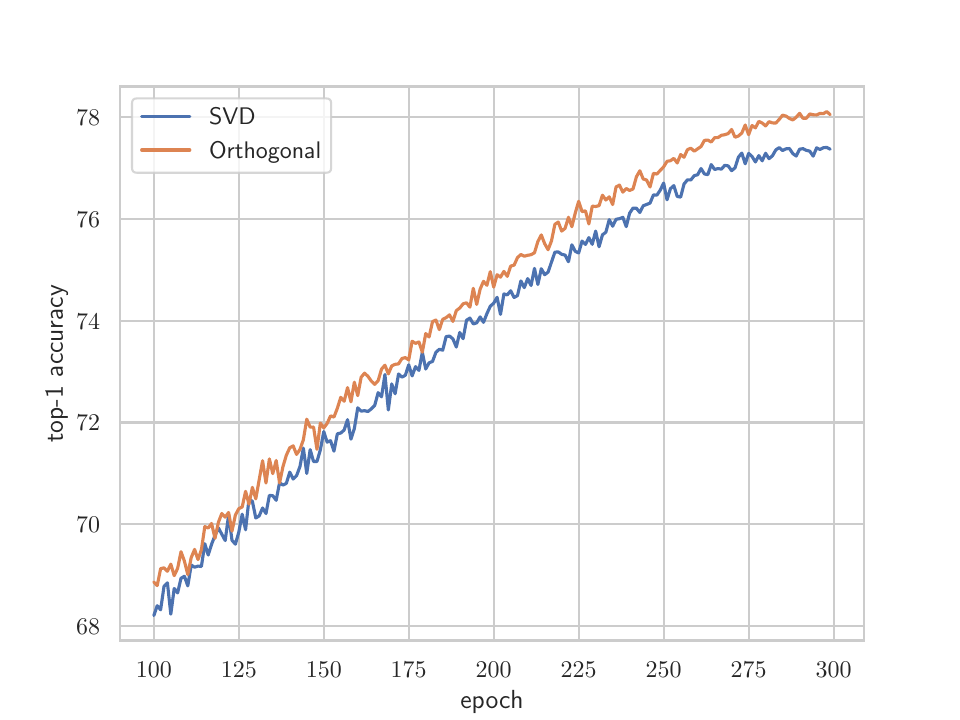}
}
\caption{Comparison between using an orthogonal projection to using a truncated SVD of the teacher features. We observe consistent improvement in performance and convergence, while being much more computationally efficient.}
\label{fig:projector_ablation_svd_vs_orthog}
\end{figure}

We compare our method to a handcrafted projection. For this projection we match the student features with a truncated SVD decomposition of the teacher features. In this way the student will align with the principle components of the teacher. The results are shown in figure \ref{fig:projector_ablation_svd_vs_orthog} and although some good performance is achieved, it falls on achieving the top-end accuracy attained by an orthogonal projection. We expect this drop in performance is likely a consequence of the improved gradient flow when performing the loss in the larger teacher space and also that the smaller principle components are indeed contributing to the discriminative power of the learned representation. Furthermore, computing the SVD is much more computationally expensive due to the expensive decomposition required for each batch.

\begin{table*}[t]
    \centering
    \footnotesize
    \setlength\tabcolsep{5pt}
\begin{tabular}{cc|cccccc}
    \toprule
    \belowrulesepcolor{mygrey}
    \rowcolor{mygrey} Distillation & Teacher & vgg13 & \cellcolor{mygrey} ResNet50 & ResNet50 & resnet32x4 & resnet32x4 & WRN-40-2 \\
    \rowcolor{mygrey}  Mechanism & Student & MobileNetV2 & MobileNetV2 & vgg8 & ShuffleNetV1 & ShuffleNetV2 & ShuffleNetV1 \\
    \aboverulesepcolor{mygrey}
    \midrule
     & Teacher & 74.64 & 79.34 & 79.34 & 79.42 & 79.42 & 75.61 \\ 
     & Student & 64.60 & 64.60 & 70.36 & 70.50 & 71.82 & 70.50 \\ 
    \midrule
    \multirow{2}{*}{Logit} & KD~\cite{Hinton2015DistillingNetwork} & 67.37 & 67.35 & \textbf{73.81} & 74.07 & 74.45 & 74.83\\
    & DKD~\cite{Zhao2022DecoupledDistillation} & \textbf{69.71} & \textbf{70.35} & - & \textbf{76.45} & \textbf{77.07} & \textbf{76.70} \\
    
    \midrule
    
    \multirow{5}{*}{Intermediate} & FitNet~\cite{Romero2015FitNets:Nets} & 64.14  & 63.16  & 70.69  & 73.59  & 73.54  & 73.73  \\
    & AT~\cite{Zagoruyko2019PayingTransfer} & 59.40  & 58.58  & 71.84  & 71.73  & 72.73  & 73.32  \\
    & NST~\cite{Huang2017LikeTransfer} & 58.16  & 64.96  & 71.28  & 74.12  & 74.68  & 74.89 \\
    & SP~\cite{Tung2019Similarity-preservingDistillation} & 66.30  & 68.08  & \textbf{73.34}  & 73.48  & 74.56  & 74.52 \\
    & ReviewKD~\cite{Chen2021DistillingReview} & \textbf{70.37} & \textbf{69.89} & - & \textbf{77.45} & \textbf{77.78}  & \textbf{77.14} \\
    
    \midrule
    
    \multirow{8}{*}{Representation} & CC~\cite{Peng2019CorrelationDistillation} & 64.86  & 65.43  & 70.25  & 71.14  & 71.29  & 71.38 \\
    & RKD~\cite{Park2019RelationalDistillation} & 64.52  & 64.43  & 71.50  & 72.28  & 73.21  & 72.21 \\
    & PKT~\cite{Passalis2018LearningTransfer} & 67.13  & 66.52  & 73.01  & 74.10  & 74.69  & 73.89  \\

    & CRD~\cite{Tian2019ContrastiveDistillation} & 69.94  & 69.54  & 74.58  & 75.12  & 76.05  & 76.27  \\
    
    & WCoRD~\cite{Chen2020WassersteinDistillation} & 70.02  & 70.12  & 74.68  & 75.77  & 76.48  & 76.68 \\

    \aboverulesepcolor{myblue}
    \rowcolor{myblue} & $V_k$D & \textbf{70.11}  & \textbf{70.65} & \textbf{74.95} & \textbf{77.05} & \textbf{77.51} & \textbf{77.19} \\

    \rowcolor{myblue} & $\Delta$ & {\color{tickgreen} \textbf{+2.74}} & {\color{tickgreen} \textbf{+3.30}} & {\color{tickgreen} \textbf{+1.14}} & {\color{tickgreen} \textbf{+2.98}} & {\color{tickgreen} \textbf{+3.06}} & {\color{tickgreen} \textbf{+2.36}} \\
    \aboverulesepcolor{myblue}
    
    \bottomrule
\end{tabular}
    \caption{CIFAR-100 test \textit{accuracy} (\%) of student networks trained with a number of distillation methods. The best results for each distillation mechanism are highlighted in \textbf{bold}. $\Delta$ represents the performance improvement over classical KD. Representation is used here to describe the features directly before the final classifier.}
    \label{table:cifar100_classif}
\end{table*}

\subsection{Experiments on CIFAR100}
CRD~\cite{Tian2019ContrastiveDistillation} provides an easy benchmark for most distillation methods. However, the results on this benchmark have become increasingly saturated, where many methods are even reporting better student performance than the teacher. This situation alone raises questions on whether the improvement is down to an improved knowledge distillation, or simply through the introduction of implicit model regularisation. Despite these limitations, we do provide some results on this CIFAR100 benchmark in table \ref{table:cifar100_classif}. Here we observe competitive performance to previous state-of-the-art on a few of the challenging cross-architecture settings. 

Although ReviewKD~\cite{Chen2021DistillingReview} achieves very strong performance on this benchmark, its application is limited to the CNN $\rightarrow$ CNN settings. Furthermore, it requires many additional trainable parameters and has a much larger memory overhead since the intermediate representations are needed to compute the loss.

\subsection{Implementation details}
\begin{figure}[H]
\centering
\resizebox{.9\linewidth}{!}{%
\includegraphics{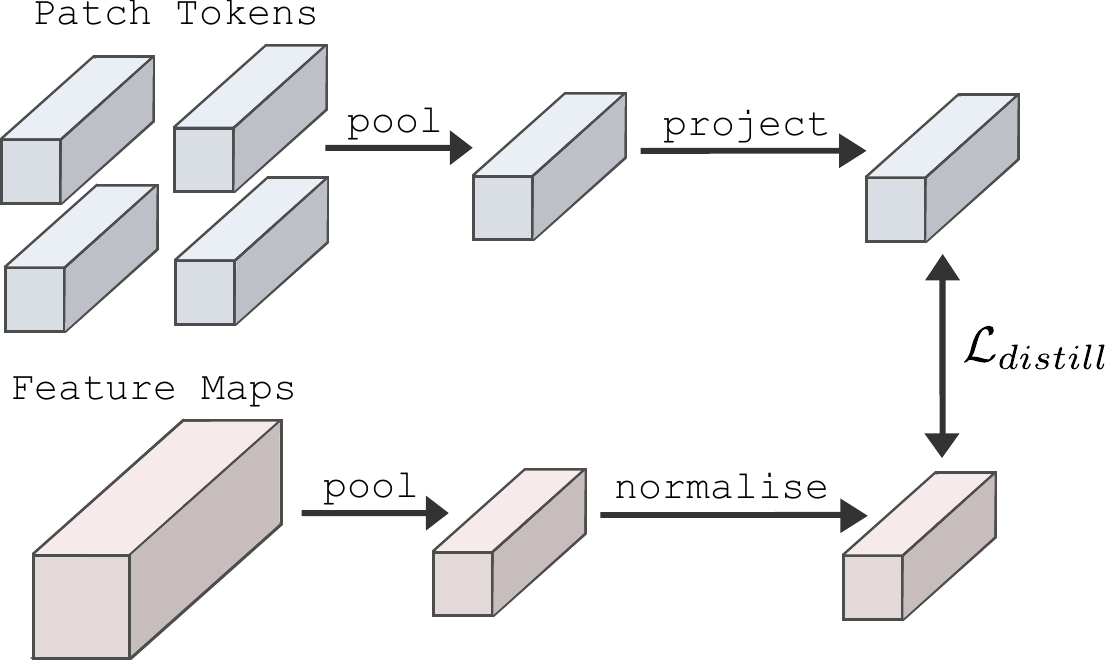}
}
\caption{Patch token distillation between transformer and CNN models. We replace the projection with an orthogonal projection, while for the normalisation we either use layer norm or iterative whitening. For the pooling we adopt a simple global average.}
\label{fig:patch_token_distillation}
\end{figure}

\paragraph{Patch token distillation} is used to distill from or to a transformer based model and can be seen in Fig. \ref{fig:patch_token_distillation}. This method was adopted since adding more distillation tokens~\cite{touvron2022deit} would introduce additional trainable parameters. Using a pooling strategy over the patch tokens proved to be very simple and effective.

\section{Qualitative results for image generation}
\begin{figure}[H]
\centering
\resizebox{.75\linewidth}{!}{%
\includegraphics{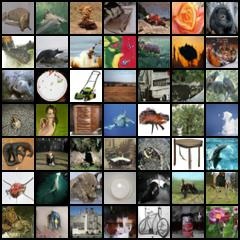}
}
\caption{Example images generated using our $V_kD$ distilled BigGAN model using the CIFAR-100 training dataset.}
\label{fig:cifar100_generated_examples}
\end{figure}

Fig. \ref{fig:cifar100_generated_examples} shows some example images generated using the $V_kD$ distilled BigGAN model trained on CIFAR100. The results demonstrate a very diverse generation of images, while preserving a lot of structural object information.

\subsection{Computational overhead} 
As shown in Tab. \ref{table:timing}, during training, we only observe a small increase in latency but almost no increase in GPU memory. Most other methods come with a significant increase in memory, while also incurring additional training time overheads. There is also no inference overhead since we throw away the orthogonal projection after training.

\begin{table}[H]
    \centering
    \footnotesize
    \begin{tabular}{lcc}
    \toprule
    \belowrulesepcolor{mygrey}
    \rowcolor{mygrey} Method & Time (s) & Memory (GiB) \\ 
    \aboverulesepcolor{mygrey}

    \midrule
    Linear & 0.72 $\pm$ 0.01 & 17.67 \\
    Orthogonal & 0.98 $\pm$ 0.02 & 17.82 \\
    \midrule
    RKD & 0.76 $\pm$ 0.01 & 19.60 \\
    VID & 0.93 $\pm$ 0.01 & 25.95 \\
    ICKD & 0.78 $\pm$ 0.02 & 19.97 \\
    CRD & 0.80 $\pm$ 0.02 & 21.46 \\
    \bottomrule

\end{tabular}
    \caption{Timing and memory of different methods in ImageNet. We distill to a DeiT-S with an effective batch size of 1024 on ImageNet using 2 NVIDIA V100 GPUs.} 
    \label{table:timing}
\end{table}

\subsection{Quantitative diversity metrics}
We provide an evaluation metric on the feature diversity in Tab. \ref{table:kdgan_recall}. We generated 5k images from each model trained with or without our whitening and also with layer normalization. We see that whitening does encourage image diversity while also improves the FID and IS realism scores.  

\begin{table}[H]
    \centering
    \footnotesize
    \begin{tabular}{clccc}
    \toprule
    \belowrulesepcolor{mygrey}
    \rowcolor{mygrey} Dataset & Metric & No Norm & Layer Norm & Whitening \\ 
    \aboverulesepcolor{mygrey}
    \midrule
    & Precision & 0.87 & 0.85 & \textbf{0.88} \\
    & Recall & 0.72 & 0.69 & \textbf{0.74} \\
    \multirow{-4}{*}{C10} & Density & \textbf{1.14} & 1.00 & 1.12 \\
    & Coverage & 0.95 & 0.95 & \textbf{0.96} \\
    \midrule
    & Precision & \textbf{0.88} & 0.85 & \textbf{0.88} \\
    & Recall & 0.04 & 0.00 & \textbf{0.71} \\
    \multirow{-4}{*}{C100} & Density & 1.07 & 0.86 & \textbf{1.17} \\
    & Coverage & 0.56 & 0.15 & \textbf{0.96} \\
    \bottomrule
\end{tabular}
    \caption{Evaluation for the fidelity and diversity of the distilled student models. Although we observe only a small improvement in recall when whitening the target teacher features for CIFAR10, we find it is critical in avoiding mode collapse for the more diverse CIFAR100 dataset.}
    \label{table:kdgan_recall}
\end{table}

\subsection{Normalisation improves convergence}
\begin{figure}[H]
\centering
\resizebox{.9\linewidth}{!}{%
\includegraphics{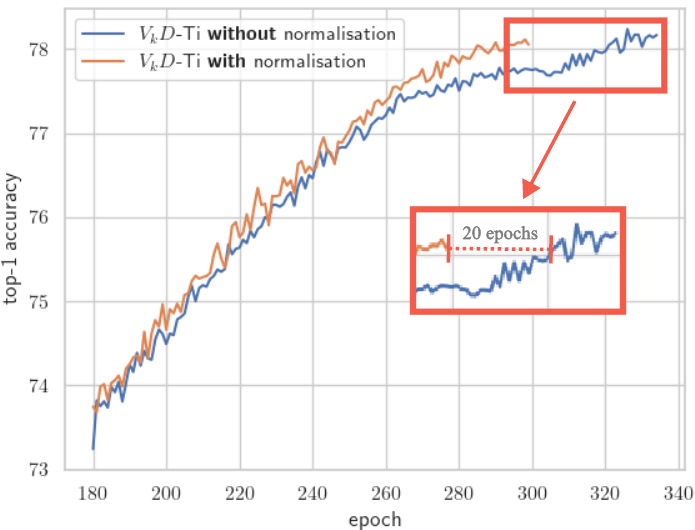}
}
\caption{Normalisation improves convergence for discriminative tasks by improving the robustness of the loss to small/irrelevant perturbations in the input image.}
\label{fig:normalisation_improves_convergence}
\end{figure}

Fig. \ref{fig:normalisation_improves_convergence} 
shows the evaluation results after each epoch of training. Here we confirm that the normalisation step does improve the model convergence. We find that simply extending the un-normalised training pipeline is enough to recover the drop in accuracy.

\subsection{Further analysis}
In this section we provide the complete derivations to supplement the illustrative analysis in the main manuscript.

\paragraph{Orthogonality of $exp(\bf{W})$.}
If $\mathbf{W}$ is a skew-symmetric matrix, then it admits the property $\mathbf{W}^T = -\mathbf{W}$. Its matrix-exponential is then given by:

\begin{equation}
    \exp(\mathbf{W}) = \mathbf{I} + \mathbf{W} + \frac{\mathbf{W}^2}{2!} + \frac{\mathbf{W}^3}{3!}\dots
\end{equation}

\noindent Since $\mathbf{W}$ is skew-symmetric, the transpose of this exponential is given as follows:

\begin{align}
    \exp(\mathbf{W})^T &= \mathbf{I} - \mathbf{W} + \frac{\mathbf{W}^2}{2!} - \frac{\mathbf{W}^3}{3!}\dots \\
    &= \exp(-\mathbf{W})
\end{align}

\noindent Thus $\exp(\mathbf{W})\exp(\mathbf{W})^T = \exp(\mathbf{W})\exp(-\mathbf{W}) = \mathbf{I}$, which confirms that $\exp(\mathbf{W})$ is indeed orthogonal.

\paragraph{Whitening and feature diversity}
In this section we provide a more thorough investigation into the connection between the use of whitening and feature diversity. Our loss is simply the pair-wise distance between the student and teacher features. Through simple algebraic manipulation, we can re-express this loss as follows:

\begin{small}
\begin{align}
    \label{eqn:ldistillfull}
    \mathcal{L}_{distill} &= \lpnorm{\mathbf{Z}^s - \mathbf{Z}^t}^2 \nonumber \\
    &= \sum_{i \neq j}\abs{\mathbf{Z}^s_{:,j} - \mathbf{Z}^t_{:,i} - \mathbf{Z}^t_{:,j} + \mathbf{Z}^t_{:,i}}^2 \nonumber \\
    &= \sum_{i \neq j}\abs{\mathbf{Z}^s_{:,j} - \mathbf{Z}^t_{:,i}}^2 + \abs{\mathbf{Z}^t_{:,j} + \mathbf{Z}^t_{:,i}}^2 \\
    & \;\;\;\;\;\;\;\;\;\;\;\;\;\;\;\;\;\;\;\;\;\;\;\;\;\;\;\;\;\; -2\braket{\mathbf{Z}^s_{:,j} - \mathbf{Z}^t_{:,i}, \mathbf{Z}^t_{:,j} + \mathbf{Z}^t_{:,i}}, \nonumber
\end{align}
\end{small}

\noindent where $\mathbf{Z}^t$ is whitened such that $(\mathbf{Z}^t)^T(\mathbf{Z}^t) = \mathbf{I}$. This means that the magnitude of each feature will be equal to one and the dot product between the different features within a batch will be zero. We can express these two properties as follows:

\vspace{-1em}
\begin{align}
    &\textit{unit length\;:\;} &\lpnorm{\mathbf{Z}^t_{:, i}}^2 &= 1, \; \nonumber \\ 
    &\textit{decorrelated\;:\;} &\braket{\mathbf{Z}^t_{:, j},\mathbf{Z}^t_{:, i}} &= 0
\end{align}

Substituting into the second term of equation \ref{eqn:ldistillfull} leads to the following simplification:

\begin{small}
\begin{align}
   &\mathcal{L}_{distill} = \sum_{i \neq j}\lpnorm{\mathbf{Z}^s_{:,j} - \mathbf{Z}^t_{:,i}}^2 + 2 - 2\braket{\mathbf{Z}^s_{:,j} - \mathbf{Z}^t_{:,i}, \mathbf{Z}^t_{:,j} + \mathbf{Z}^t_{:,i}} \nonumber
\end{align}
\end{small}

Using the Cauchy-Schwartz inequality, we can find a lower bound on this loss.

\begin{small}
\begin{align}
    \mathcal{L}_{distill} &\geq \sum_{i \neq j}\lpnorm{\mathbf{Z}^s_{:,j} - \mathbf{Z}^t_{:,i}}^2 + 2 - 2\sqrt{\lpnorm{\mathbf{Z}^s_{:,j} - \mathbf{Z}^t_{:,i}}^2\lpnorm{\mathbf{Z}^t_{:,j} + \mathbf{Z}^t_{:,i}}^2} \nonumber \\
    &\geq \sum_{i \neq j}\lpnorm{\mathbf{Z}^s_{:,j} - \mathbf{Z}^t_{:,i}}^2 + 2 - 2\lpnorm{\mathbf{Z}^s_{:,j} - \mathbf{Z}^t_{:,i}}^2\lpnorm{\mathbf{Z}^t_{:,j} + \mathbf{Z}^t_{:,i}}^2 \nonumber \\ 
    &= \sum_{i \neq j}\lpnorm{\mathbf{Z}^s_{:,j} - \mathbf{Z}^t_{:,i}}^2 + 2 -4\lpnorm{\mathbf{Z}^s_{:,j} - \mathbf{Z}^t_{:,i}}^2 \nonumber \\
    &= \sum_{i \neq j}2-3\lpnorm{\mathbf{Z}^s_{:,j} - \mathbf{Z}^t_{:,i}}^2 \nonumber \\
    &= \text{const} - 3\sum_{i \neq j}\underbrace{\lpnorm{\mathbf{Z}^s_{:,j} - \mathbf{Z}^t_{:,i}}^2}_{\mathbf{C}^2}
    \label{eqn:bound}
\end{align}
\end{small} 

This bound is minimised when the distance between each $i \neq j$ student and teacher feature is maximised. Similarly, the $L2$ loss itself in equation \ref{eqn:ldistillfull} minimises the pair-wise distance between features. These two results show that the whitening operation jointly minimises the pairwise similarity, while also maximising an upper bound for the cross feature diversity.

\end{document}